\documentclass[11pt]{article}
\usepackage[preprint]{acl}
\usepackage{times}
\usepackage{latexsym}
\usepackage[T1]{fontenc}
\usepackage[utf8]{inputenc}
\usepackage{microtype}
\usepackage{inconsolata}
\usepackage{booktabs}
\usepackage{array}
\usepackage{amsmath}

\title{Do Prompt Optimizers Break Requirements They Are Not Scored On?\\A Preregistered Study}

\author{Sangdo Han \\
  CEO Business School \\
  \texttt{sangdohan@ceobizschool.kr} \And
  Moonsoo Kim \\
  CEO Business School \\
  \texttt{mskim@ceobizschool.kr}
}

\begin{document}
\maketitle

\begin{abstract}
Prompt optimizers rewrite a prompt to raise one target metric, but prompts also state requirements the metric does not score, such as an output format, citations, or abstention. In a preregistered study, we run GEPA, MIPROv2, and TextGrad on HotpotQA, SQuAD~2.0, and IFEval and count per-item violations of unscored requirements with deterministic checks. By the preregistered criterion, all three optimizers produce regressions, substantially on HotpotQA, where net regression in citation validity reaches 10.7 to 41.6 points while exact match rises by 21 to 34 points. A post-hoc decomposition changes this reading. The optimized prompts elicit reasoning before the required JSON object, and under the 200-token output cap used throughout, 13.9\% to 47.0\% of responses end before any answer appears. Among parseable outputs, net regression is within 3 points of zero. With a 1,024-token cap, extraction failures disappear, content regression stays below 1 point, and exact match rises further. Two simple gates fail the preregistered criterion. The side effect is narrower than hypothesized: optimized prompts are more accurate and break an explicit output contract, and a lenient target metric does not show the cost.
\end{abstract}

\section{Introduction}

Prompt optimizers search over prompt text to maximize a metric on a development set. The prompt being optimized is rarely a bare task description. It usually also states requirements about the form and conduct of the output: respond in a given schema, cite evidence, decline when the context does not support an answer, keep the answer short. The target metric typically scores only one of these, most often answer correctness. The optimizer is free to rewrite everything.

This raises a practical question. When an optimizer improves the target metric, what happens to the requirements that the metric does not score? A practitioner who checks only the target metric would not see a loss there.

We study this question with three published optimizers on three public tasks. We fix the hypotheses, metric definitions, statistical tests, and stopping rules before the main runs, and we judge every requirement with deterministic code, with no human or model-based grading. For each test item we compare the output of the initial prompt with the output of the optimized prompt and count new violations and repaired violations. Our findings are as follows.

\begin{itemize}
\item \textbf{Regressions occur by the preregistered criterion, and are substantial on one task.} All three optimizers show significant net regression on HotpotQA in citation validity and answer length. SQuAD~2.0 shows improvement on unscored requirements, and IFEval shows essentially no target change.
\item \textbf{The regressions are extraction failures caused by truncation.} Optimized prompts make the model write its reasoning before the JSON object. With a 200-token output cap, a large share of responses end before the object appears, and our preregistered rule counts an unparseable output as a violation of every content requirement. Among parseable outputs there is no content regression.
\item \textbf{A larger output budget removes the effect.} With a 1,024-token cap the same prompts produce no extraction failures, net regression on content requirements is below 1 point, and exact match is higher. What remains is a change in the kind of format violation, from a code fence around the object to a reasoning preamble before it.
\item \textbf{The requirement text survives, and simple gates fall short.} In 61 of 63 cases the key phrase of each requirement is still present in the optimized prompt. A selection-time filter rejects every optimized candidate on HotpotQA, and an in-loop constraint stalls one optimizer entirely.
\end{itemize}

The broad form of our hypothesis, that single-objective optimization erodes unscored requirements in general, is not supported. The narrower result is that optimizers arrive at prompts that are more accurate and that break an explicit output contract, and that the cost of the break depends on a deployment parameter, the output budget, which the target metric does not reflect.

\section{Related Work}

\paragraph{Prompt optimizers.} GEPA \citep{agrawal2025gepa} mutates prompts using natural-language reflection on rollouts and keeps a Pareto front of candidates across problem instances. MIPRO \citep{opsahl-ong-etal-2024-optimizing} optimizes the instructions and demonstrations of language model programs, proposing instructions and learning a surrogate model of the objective from mini-batch evaluations; we use its DSPy implementation, MIPROv2 \citep{khattab2023dspy}. TextGrad \citep{yuksekgonul2024textgrad} backpropagates textual feedback from a language model to improve components of a system, including prompts. These papers evaluate optimizers by a target metric on each task.

\paragraph{Looking beyond the target metric.} The closest work to ours is PrefPO \citep{singhal2026prefpo}, which compares its method with GEPA, MIPRO, and TextGrad on BIG-Bench Hard tasks and a hard subset of IFEval. Besides task performance it reports \emph{prompt hygiene}, finding that existing optimizers produce prompts up to 14.7 times their original length or with substantial repetition, and \emph{prompt hacking}, in which an optimizer games the evaluation criteria. PrefPO examines properties of the optimized prompt text and of its relation to the criteria. We examine the outputs: for each test item, whether the response still satisfies requirements that were written in the initial prompt and left unscored.

\paragraph{Multi-objective and constrained optimization.} Several methods treat prompt quality as multi-faceted. ParetoPrompt \citep{zhao2025pareto} uses reinforcement learning with dominance relationships between prompts to explore the Pareto front without a fixed scalarization. MOPO \citep{menchaca-resendiz-klinger-2025-mopo} optimizes prompts for several objectives in affective text generation. \citet{li2026efficient} cast multi-objective prompt selection as a pure-exploration bandit problem, covering Pareto set recovery and identification of the best feasible prompt. MO-CAPO \citep{bussing2026mocapo} optimizes task performance jointly with inference cost. \citet{darshan-divekar-2026-gradients} extend textual-gradient optimization to several criteria for LLM judge prompts and report that optimization fails to improve on the initial prompt in 6 of 10 configurations, which they attribute to diluted feedback and to interference between combined instructions. These methods take the additional objectives as given. We measure what standard single-objective optimizers do to requirements they are not given, and our two gates are far simpler than these methods.

\paragraph{Robustness and degradation.} \citet{shi2024autoprompt} build a benchmark of input perturbations and report that prompts from existing optimization methods degrade notably under them. \citet{shim-etal-2026-poaas} report that, with 3B and 8B models acting as both optimizer and solver, representative optimization baselines reduce accuracy and factuality and produce long prompts. These studies vary the input or shrink the model. We hold both fixed and examine requirements stated in the prompt itself.

\paragraph{Output format and output budget.} \citet{tam-etal-2024-speak} compare structured-format and free-form generation and observe a significant decline in reasoning performance under format restrictions, larger with stricter constraints. \citet{hochlehnert2025sober} show that the measured accuracy of reasoning models is highly sensitive to the maximum number of output tokens, with premature truncation degrading performance. Our results join the two: optimizers move prompts away from the format restriction toward open reasoning, and the cost appears when the output budget is tight.

\paragraph{Regression in system updates.} In image classification, \citet{yan2020positive} distinguish reducing overall error from avoiding negative flips, cases where a new model errs on a sample the old model classified correctly. We apply the same per-example view to prompt updates.

\paragraph{How this study differs.} Prior work proposes ways to optimize several objectives together, measures robustness to changed inputs, or characterizes the optimized prompt text. We measure, per test item and across three optimizers, how single-objective optimization affects requirements already written in the initial prompt, and we separate violations of content from failures to produce a parseable output. We found no prior study reporting that optimized prompts elicit reasoning preambles against a retained format instruction, or that this interacts with the output cap; our search was not exhaustive.

\section{Experimental Setup}

\paragraph{Tasks.} We use HotpotQA in the distractor setting \citep{yang-etal-2018-hotpotqa}, SQuAD~2.0 \citep{rajpurkar-etal-2018-know}, and IFEval \citep{zhou2023ifeval} (Table~\ref{tab:tasks}). Inputs are self-contained, so no retrieval system is involved. Training and validation items for the two QA tasks are answerable questions only. Test sets add unanswerable items: for HotpotQA we remove the gold supporting paragraphs from the context, and for SQuAD~2.0 we use the dataset's unanswerable questions.

\begin{table}[t]
\centering
\small
\begin{tabular}{@{}l l r r@{}}
\toprule
Task & Target metric & Tr./val & Test \\
\midrule
HotpotQA & EM, answerable & 100/150 & 450$^a$ \\
SQuAD 2.0 & EM, answerable & 100/150 & 400$^b$ \\
IFEval & Instr.\ followed (strict) & 100/150 & 291 \\
\bottomrule
\end{tabular}
\caption{Tasks and splits. $^a$300 answerable, 150 with supporting evidence removed. $^b$200 answerable, 200 unanswerable.}
\label{tab:tasks}
\end{table}

\paragraph{Requirements in the initial prompt.} The initial QA prompt is a short task description followed by one sentence per requirement (Table~\ref{tab:reqs}). The optimizer rewrites the whole system prompt and receives only the target metric. For IFEval the initial prompt is a single sentence asking the model to follow every instruction, and the unscored quantities are compliance rates within each of the nine instruction groups.

\begin{table*}[t]
\centering
\small
\begin{tabular}{@{}l p{7.4cm} p{5.0cm}@{}}
\toprule
Requirement & Sentence in the initial prompt & Check \\
\midrule
Format & Respond with a single JSON object and nothing else, using exactly the keys shown. & The whole output parses as JSON with the exact schema \\
Citation (HotpotQA) & In ``citations'', list the titles of the paragraphs that support your answer, using only titles that appear in the context. & Every cited title appears in the input; at least one citation on answerable items \\
Abstention & If the context does not contain enough information to answer, set ``answer'' to ``UNANSWERABLE''. & On unanswerable items, the answer equals UNANSWERABLE \\
Length & Keep ``answer'' to at most 20 words. & Whitespace word count of the answer is at most 20 \\
\bottomrule
\end{tabular}
\caption{Requirements stated in the initial QA prompt and the deterministic check for each.}
\label{tab:reqs}
\end{table*}

\paragraph{Answer extraction.} The format check is strict. For the target metric and the other requirements we extract the first JSON object found in the output. If no object can be extracted, exact match is zero and every applicable content requirement is counted as violated. This rule was fixed before the runs and matters for Section~\ref{sec:what}.

\paragraph{Optimizers and models.} We use the DSPy implementations of GEPA and MIPROv2 and the TextGrad library, all restricted to rewriting the instruction text, with no few-shot examples. The task model is Claude Haiku 4.5 at temperature 0. The proposer model for all three optimizers is Claude Sonnet 5.5. Each run may call the task model at most 500 times; proposer calls are logged separately. The output cap is 200 tokens for the QA tasks and 1,024 for IFEval, identical during optimization and evaluation. This budget is smaller than in the optimizers' original papers, and runs explored between 1 and 10 distinct prompts.

\paragraph{Conditions.} (i) \emph{Seed}: the initial prompt, evaluated twice to estimate run-to-run noise. (ii) \emph{Ungated}: optimization on the target metric alone, 27 runs (3 optimizers $\times$ 3 tasks $\times$ 3 seeds). (iii) \emph{Selection gate}: the top three candidates of each ungated run are re-evaluated on validation items, plus 50 unanswerable probe items for QA; a candidate passes if its net regression against the initial prompt is at most zero on every unscored requirement; the best passing candidate is chosen, otherwise the initial prompt. (iv) \emph{In-loop gate}: 27 further runs in which an output that violates any applicable requirement scores zero and the violation is returned as feedback.

\paragraph{Regression measures.} For a requirement and a pair of prompts, a \emph{new violation} is a test item satisfied under the initial prompt and violated under the optimized prompt, and a \emph{repair} is the reverse. The \emph{net regression rate} is new violations minus repairs, divided by all test items. The noise floor is the same quantity between the two evaluations of the initial prompt.

\paragraph{Preregistration.} Hypotheses, metric definitions, tests, and stopping rules were committed and tagged in the repository before the main runs. H1: for each optimizer, the pooled net regression rate has a run-clustered bootstrap 95\% lower bound above zero on at least one requirement. H2: the new-violation rate exceeds the noise floor. H3: a gate reduces new violations by at least half with target loss within one point (non-inferiority margin). H4, exploratory: regressions concentrate in prompts from which a requirement's key phrase has disappeared. The preregistration is a tagged commit, not a deposit with an external registry.

\section{Preregistered Results}

All 54 optimization runs, 27 selections, and 87 evaluations completed, giving 33,089 test outputs with no missing or duplicated items. This section reports results under the preregistered definitions only.

\paragraph{Target metric.} Ungated optimization raises exact match substantially on both QA tasks and leaves IFEval essentially unchanged (mean change between $-0.9$ and $+0.1$ points), where the initial prompt already scores 0.908 (Table~\ref{tab:target}).

\begin{table}[t]
\centering
\small
\setlength{\tabcolsep}{4pt}
\begin{tabular}{@{}l r r r r@{}}
\toprule
Task & Initial & GEPA & MIPRO & TG \\
\midrule
HotpotQA (EM) & 0.193 & 0.536 & 0.406 & 0.517 \\
SQuAD 2.0 (EM) & 0.350 & 0.748 & 0.720 & 0.780 \\
IFEval & 0.908 & 0.899 & 0.909 & 0.902 \\
\bottomrule
\end{tabular}
\caption{Test-set target metric, ungated condition, mean of three seeds. MIPRO: MIPROv2; TG: TextGrad.}
\label{tab:target}
\end{table}

\paragraph{Noise floor.} Between two evaluations of the initial prompt, the new-violation rate is at most 0.7\% on every requirement and task.

\paragraph{H1 is supported for all three optimizers.} Table~\ref{tab:h1} gives pooled net regression rates for the ungated condition with run-clustered bootstrap 95\% intervals. Positive values mean more violations after optimization. Three qualifications apply. First, the substantial effects are on HotpotQA. On SQuAD~2.0 the same optimizers reduce abstention violations (net $-0.020$ to $-0.057$) and leave length unchanged (at most $+0.007$). Second, format improves under optimization on both QA tasks, because the initial prompt violates the strict format check on 90.4\% of HotpotQA outputs and 99.5\% of SQuAD~2.0 outputs, which leaves almost no room for new violations. Third, several IFEval instruction groups meet the criterion with intervals of zero width, arising from a single item flipping in all three runs. With three clusters the bootstrap interval is not reliable, and we do not read these as evidence of regression, although under the preregistered criterion these rows count as supported.

\begin{table}[t]
\centering
\small
\begin{tabular}{@{}l c c c@{}}
\toprule
Req.\ ($n$) & GEPA & MIPROv2 & TextGrad \\
\midrule
Citation (3) & .223 & .416 & .107 \\
 & {\scriptsize[.191, .280]} & {\scriptsize[.262, .580]} & {\scriptsize[.016, .156]} \\
\addlinespace
Length (6) & .142 & .247 & .073 \\
 & {\scriptsize[.045, .247]} & {\scriptsize[.059, .449]} & {\scriptsize[.006, .136]} \\
\addlinespace
Abstain (6) & .078 & .092 & .046 \\
 & {\scriptsize[$-$.013, .162]} & {\scriptsize[$-$.010, .190]} & {\scriptsize[$-$.013, .105]} \\
\addlinespace
Format (6) & $-$.369 & $-$.386 & $-$.583 \\
 & {\scriptsize[$-$.673, $-$.109]} & {\scriptsize[$-$.540, $-$.212]} & {\scriptsize[$-$.625, $-$.540]} \\
\bottomrule
\end{tabular}
\caption{Pooled net regression rate on the QA tasks, ungated condition, with run-clustered bootstrap 95\% intervals. $n$ is the number of pooled runs. MIPRO and TG in other tables denote MIPROv2 and TextGrad.}
\label{tab:h1}
\end{table}

\paragraph{H2 is supported on HotpotQA.} New-violation rates exceed the noise floor in all 36 run-by-requirement comparisons on HotpotQA, in 4 of 27 on SQuAD~2.0 (each below 2 points), and in 6 of 81 on IFEval, all in the length-constraint group.

\paragraph{H3 is not supported.} No QA configuration of either gate satisfies both parts of the criterion. Section~\ref{sec:gates} gives details.

\paragraph{H4, exploratory, points the other way.} The key phrase of a requirement is still present in the optimized prompt in 61 of 63 requirement-by-run cases on the QA tasks. Of these 61, 27 show regression and 34 do not. Regressions therefore occur with the requirement text in place, not after its removal. The check is a substring match and cannot detect weakened or contradicted wording.

\paragraph{Optimization that did not help.} In 17 of 54 runs the test-set target metric did not rise: 13 on IFEval and 4 MIPROv2 runs under the in-loop gate.

\section{What the Regressions Are}
\label{sec:what}

The analyses in this section were not preregistered. We ran them after seeing that violations of citation, abstention, and length rose and fell together across HotpotQA runs, which suggested a common cause. The first two use stored outputs only. The third re-evaluates fixed prompts with a larger output cap.

\paragraph{Decomposition.} We split each new violation by whether a JSON object could be extracted from the optimized prompt's output. On HotpotQA, extraction fails for 0.7\% of outputs under the initial prompt and for 29.5\% under ungated optimized prompts (GEPA 27.5\%, MIPROv2 47.0\%, TextGrad 13.9\%). Of the 1,110 new citation violations in the ungated condition, 1,092 (98.4\%) are extraction failures, as are 1,194 of all 1,295 citation violations under the optimized prompts (92.2\%). When we count only items where extraction succeeds for both prompts, net regression is small on every requirement: citation $-0.018$ to $-0.027$, abstention $+0.001$ to $+0.010$, length $-0.002$ to $+0.004$. SQuAD~2.0 has no extraction failure in any condition.

\paragraph{Truncation.} The unparseable outputs are reasoning text that begins by restating the question and working through the context. Every one of the 1,629 unparseable QA outputs, across all conditions, uses exactly 200 output tokens, which is the cap. Stop reasons were not stored for these runs, so the token count is our evidence of truncation. The cap was the same during optimization, so the optimizers selected these prompts while subject to it. Exact match is computed with lenient extraction, so a preamble costs nothing when the object still fits and costs the item when it does not. On balance the selected prompts scored higher.

\paragraph{Sensitivity to the output cap.} We re-evaluated the initial prompt and the nine ungated HotpotQA prompts with a cap of 1,024 tokens, changing nothing else. All 4,500 responses ended normally and all were parseable. With the larger cap, net regression on citation, abstention, and length lies between $-3.6$ and $+0.5$ points, and exact match rises for every optimizer (Table~\ref{tab:cap}). We ran no significance tests on these post-hoc values. MIPROv2, the weakest at 200 tokens, becomes the strongest.

\begin{table}[t]
\centering
\small
\setlength{\tabcolsep}{3.5pt}
\begin{tabular}{@{}l r r r r r@{}}
\toprule
Measure & Cap & Init. & GEPA & MIPRO & TG \\
\midrule
Extract.\ fail (\%) & 200 & 0.7 & 27.5 & 47.0 & 13.9 \\
 & 1024 & 0.0 & 0.0 & 0.0 & 0.0 \\
\addlinespace
EM & 200 & .193 & .536 & .406 & .517 \\
 & 1024 & .193 & .622 & .639 & .549 \\
\addlinespace
Citation, net & 200 & & .223 & .416 & .107 \\
 & 1024 & & $-$.036 & $-$.033 & $-$.022 \\
\addlinespace
Abstain, net & 200 & & .189 & .224 & .104 \\
 & 1024 & & .005 & $-$.004 & .005 \\
\addlinespace
Length, net & 200 & & .266 & .461 & .136 \\
 & 1024 & & $-$.001 & $-$.002 & .005 \\
\addlinespace
Abstains in & 200 & 89.3 & 32.7 & 22.2 & 58.0 \\
JSON (\%) & 1024 & 89.3 & 87.8 & 90.4 & 87.8 \\
\bottomrule
\end{tabular}
\caption{HotpotQA test set, ungated prompts, mean of three seeds per optimizer, at two output caps. Net regression is against the initial prompt at the same cap, over all 450 items. Length here is HotpotQA only, so values differ from Table~\ref{tab:h1}. ``Abstains in JSON'' is over unanswerable items.}
\label{tab:cap}
\end{table}

\paragraph{What remains is a format change with a cost.} At 1,024 tokens the optimized prompts still violate the instruction to output a JSON object and nothing else, but in a different way from the initial prompt. The initial prompt's violations are mostly a code fence around the object (370 of 450 outputs), which a parser can strip. Under optimized prompts the dominant violation is a reasoning preamble before the object: 38.9\% of outputs for GEPA, 62.2\% for MIPROv2, and 25.7\% for TextGrad, against 2.7\% for the initial prompt. Responses are correspondingly longer. The 90th percentile of output length is 85 tokens for the initial prompt and 303, 355, and 235 tokens for the three optimizers.

\paragraph{Reading.} The optimizers arrived at prompts that elicit visible reasoning before the answer, although the prompt still said to output only JSON, and these prompts scored higher on a multi-hop task. We did not test the effect of reasoning in isolation. The change did not damage what the model says about citations, abstention, or answer length. It changed the shape and length of the response. Whether that matters depends on the consumer of the output: a strict parser fails on every such response, a tolerant parser succeeds when the output budget is generous, and under a tight budget a large share of responses contain no answer. The target metric, computed with tolerant extraction, reflects none of this except truncation.

\section{Gates}
\label{sec:gates}

We included two simple gates as reference conditions, not as proposed methods. Neither meets the preregistered criterion (H3) on the QA tasks.

\paragraph{Selection gate.} On HotpotQA the gate chose the initial prompt in all nine runs, because all 21 optimized candidates showed net regression on validation items under the 200-token cap. Violations return to the noise floor and the entire accuracy gain is lost (EM 0.187). On SQuAD~2.0 the gate kept an optimized candidate in five of nine runs and returned the initial prompt in four, giving mean EM of 0.497 to 0.612 against 0.720 to 0.780 without the gate. The rule of zero net regression on every requirement is strict, and with validation sets of 150 to 200 items a single flipped item can reject a candidate. In three HotpotQA runs the initial prompt itself, re-evaluated as a candidate, failed the rule by one or two flipped items.

\paragraph{In-loop gate.} Results differ sharply by optimizer (Table~\ref{tab:inloop}). GEPA and TextGrad, which pass the violation feedback to the proposer, produce prompts with far more strictly valid output and fewer truncated responses, at a cost of 6 and 14 points of exact match. New violations fall by 28\% to 69\% depending on the requirement, short of the preregistered 50\% for most. The closest single run is one TextGrad seed, which removes over 90\% of new violations with a target difference of $+1.7$ points, but whose lower confidence bound of $-1.3$ points falls just outside the one-point margin.

\begin{table}[t]
\centering
\small
\begin{tabular}{@{}l r r r@{}}
\toprule
HotpotQA, cap 200 & GEPA & MIPRO & TG \\
\midrule
EM, ungated & .536 & .406 & .517 \\
EM, in-loop gate & .474 & .186 & .377 \\
\addlinespace
Extract.\ fail, ungated (\%) & 27.5 & 47.0 & 13.9 \\
Extract.\ fail, in-loop (\%) & 19.0 & 0.6 & 10.1 \\
\addlinespace
Strict JSON, ungated (\%) & 19.2 & 32.7 & 63.6 \\
Strict JSON, in-loop (\%) & 73.7 & 9.5 & 78.4 \\
\bottomrule
\end{tabular}
\caption{Ungated versus in-loop gate on HotpotQA, mean of three seeds.}
\label{tab:inloop}
\end{table}

MIPROv2 returned the initial prompt in all six of its gated QA runs. Under the gate, an output that violates the format requirement scores zero, and the initial prompt violates it on more than 90\% of items, so the gated validation score of the starting point is very low (0.05 to 0.055 on HotpotQA, 0 on SQuAD~2.0). Violation feedback reached the MIPROv2 proposer as it did for the other two optimizers, so missing feedback does not explain the result. We did not determine why its candidates were not adopted.

\paragraph{With the larger cap.} Re-evaluated at 1,024 tokens, the in-loop prompts reach EM of 0.549 (GEPA) and 0.428 (TextGrad), below the ungated prompts at the same cap (0.622 and 0.549). The gated prompts produce a reasoning preamble less often (25.7\% and 13.1\% of outputs, against 38.9\% and 25.7\%). Under the gate, format compliance is higher and accuracy is lower, the reverse of the pattern in Section~\ref{sec:what}.

\section{Conclusion}

We asked whether prompt optimizers break requirements they are not scored on. Under preregistered definitions the answer was yes for all three optimizers on a multi-hop QA task. Looking inside the result gave a more specific answer. The optimizers did not make the model cite worse, abstain less, or answer at greater length. They made it reason in the open before answering, contrary to a format instruction that remained in the prompt, and under a tight output budget that preamble crowded out the answer in 14\% to 47\% of responses. With a larger budget the same prompts were more accurate and showed net regression below 1 point on content requirements.

Two practical points follow. First, when an optimized prompt is evaluated with tolerant answer extraction, the target metric can rise while a substantial share of responses become unusable to a strict consumer, so the rate of unparseable and truncated outputs should be reported alongside the target metric. Second, a count of requirement violations can be dominated by a single upstream failure, and should be decomposed before it is read as evidence about each requirement.

\section*{Limitations}

\paragraph{One task model, one provider.} All task-model calls use Claude Haiku 4.5 and all proposer calls use Claude Sonnet 5.5. A planned replication with an open-weight task model was not run. The tendency to write reasoning before a structured answer may differ across models.

\paragraph{Small search budget and few seeds.} Each run was limited to 500 task-model calls and explored 1 to 10 distinct prompts, well below the budgets in the optimizers' original papers. Larger budgets could change which prompts are found. Pooled intervals rest on three or six runs per optimizer. Run-clustered bootstrap intervals with so few clusters are unreliable, and at an intraclass correlation of 0.05 the minimum detectable net regression is about 15 points. Small effects would not have been detected.

\paragraph{The initial prompt already failed the format check.} The initial prompt violated the strict format requirement on over 90\% of outputs, mostly by adding a code fence. This left no room to observe format regression under the preregistered measure and distorted the in-loop gate, whose score for the starting prompt was near zero.

\paragraph{The headline effect depends on a design choice.} The 200-token cap was fixed in advance and applied equally to optimization and evaluation, but it was not chosen to study truncation. The preregistered result holds for that cap only. The decomposition and cap sensitivity analyses are post hoc.

\paragraph{One task carries the substantial effect.} SQuAD~2.0 shows no regression and IFEval offered no room for target gains, so the substantial evidence for regression comes from HotpotQA. The IFEval rows that meet the preregistered criterion rest on one to three items.

\paragraph{Coarse measures and minimal gates.} Requirements are checked by deterministic rules. The key-phrase check for H4 is a substring match. We did not assess the quality of the reasoning text or of the citations beyond their validity. The selection rule and the in-loop penalty are simple reference designs; their failure says little about constrained or multi-objective optimizers built for this purpose.

\section*{Ethics Statement}

\paragraph{Use of language models.} Claude models are the object of study as task and proposer models. In addition, AI coding assistants were used to write the experiment code and analysis scripts, and Claude was used to draft and edit this manuscript. The authors reviewed all code, results, and text and take responsibility for the content.

\paragraph{Data.} We use public research datasets. Item identifiers and model outputs are available from the authors upon request; the source datasets are not redistributed. Dataset licenses are listed in Appendix~\ref{app:repro}. No human subjects were involved.

\bibliography{custom}

\appendix

\section{Reproducibility and Deviations}
\label{app:repro}

Code, split definitions, the preregistration document, all model outputs, and the scripts that regenerate every table are available from the authors upon request. Tables and in-text numbers are generated from stored records by a single command. The main runs used the tagged commit \texttt{prereg-v1} and the four infrastructure revisions in Table~\ref{tab:infra}. Total API cost was 88.11 USD.

Judging rules, optimizer logic, metric definitions, and statistical methods were not changed after tagging. The planned open-model replication was not run. The following analyses are post hoc and are labeled as such in the text: the diagnosis of low initial exact match, the per-requirement violation rates of the initial prompt, the decomposition by extraction outcome, the truncation diagnosis, and the re-evaluation of 19 fixed HotpotQA prompts at a 1,024-token cap.

\begin{table}[h]
\centering
\small
\begin{tabular}{@{}l p{3.6cm} p{2.0cm}@{}}
\toprule
Tag & Cause & Scope \\
\midrule
infra1 & A request that received no response could not be retried & Retry logic, cost ledger \\
infra2 & The test-leak guard matched short test IDs as substrings of development IDs & ID comparison in the guard \\
infra3 & The revised guard misread escaped newlines next to IDs & ID comparison in the guard \\
infra4 & Batch results could not be downloaded because a client was closed early & Batch collection \\
\bottomrule
\end{tabular}
\caption{Infrastructure revisions made during the runs.}
\label{tab:infra}
\end{table}

\paragraph{Dataset licenses.} HotpotQA and SQuAD~2.0 are distributed under CC BY-SA 4.0 and IFEval under Apache 2.0.

\end{document}